\documentclass[11pt]{article}
\usepackage{natbib}
\usepackage{amsmath}
\usepackage{graphicx}
\usepackage{hyperref}
\usepackage{listings}

\usepackage{subcaption} 
\usepackage{xcolor}
\usepackage{booktabs}
\usepackage{microtype}
\usepackage[margin=1in]{geometry}
\graphicspath{{./figs/}}
\usepackage{wrapfig}
\usepackage{algorithm}
\usepackage{algpseudocode}  

\title{AGI Is Coming... Right After AI Learns to Play Wordle}
\author{Sarath Shekkizhar$^*$ \hspace{2em} Romain Cosentino\thanks{Equal Contribution (random order). Salesforce.  Correspondence to: \texttt{\{sshekkizhar, rcosentino\}@salesforce.com}}
}
\date{}

\begin{document}

\maketitle

\begin{abstract}
This paper investigates multimodal agents, in particular,  OpenAI's Computer-User Agent (CUA), trained to control and complete tasks through a standard computer interface, similar to humans. We evaluated the agent's performance on the New York Times Wordle game to elicit model behaviors and identify shortcomings. Our findings revealed a significant discrepancy in the model's ability to recognize colors correctly depending on the context. The model had a $5.36\%$ success rate over several hundred runs across a week of Wordle. Despite the immense enthusiasm surrounding AI agents and their potential to usher in Artificial General Intelligence (AGI), our findings reinforce the fact that even simple tasks present substantial challenges for  today’s frontier AI models. We conclude with a discussion of the potential underlying causes, implications for future development, and research directions to improve these AI systems.
\end{abstract}

\section{Introduction}
Recent advances in multimodal (image, audio, text) models \citep{chen2025janus, pixtral2024, llama2024, openai2025cua, anthropic2024cua} have generated unprecedented excitement about the potential emergence of Artificial General Intelligence (AGI). The pace of development in language understanding, image generation, and agent-based systems has led many researchers and technology leaders to believe that AGI is right around the corner. While these systems display increasingly advanced, human-like capabilities, the lack of clearly defined limitations makes claims of such technological singularity unsubstantiated. This distinction is crucial for accurately assessing our current position and determining what is still required to achieve true agentic capabilities.

OpenAI's Computer-User Agent (CUA) \citep{openai2025cua} represents one such advancement that has refueled this optimism. CUA models enables AI systems to perceive and interact with computer interfaces through raw pixel processing, reasoning,  and programmatically controlling mouse and keyboard. OpenAI describes the model as:

\begin{quote}
``CUA combines GPT-4o's vision capabilities with advanced reasoning through reinforcement learning. CUA is trained to interact with graphical user interfaces (GUIs)—the buttons, menus, and text fields people see on a screen—just as humans do. This gives it the flexibility to perform digital tasks without using OS-or web-specific APIs.''
\end{quote}

Under the hood, the computer using agent takes as input input images to understand screen content (via screenshots), text instructions to identify next steps, and computer tools (click, type, scroll, etc...) to perform actions on the computer. This operational flow of perception, reasoning, and action to navigate digital environments is performed iteratively until a given task is completed or requires further input from user. The CUA model is the state-of-the-art model, at the time of this study, with impressive benchmark reported in \citet{openai2025cua}: $38.1\%$ success rate on OSWorld \citep{xie2024osworld} for computer use tasks, $58.1\%$ on WebArena \citep{zhou2023webarena}, and $87\%$ on WebVoyager \citep{he2024webvoyager} for web-based tasks.

In this work, we tested CUA on a simple word puzzle, the Wordle. This game presents an interesting scenario where there is a lot of information about the game available on the internet, which the model was likely trained on, while the gameplay itself is less likely to have been integrated into the model’s training process. This thought process led us to believe that the game will be an excellent setting for evaluating a multimodal agent's perceptual and reasoning capabilities.

\begin{figure}[htbp]
\centering
\includegraphics[width=0.34\textwidth]{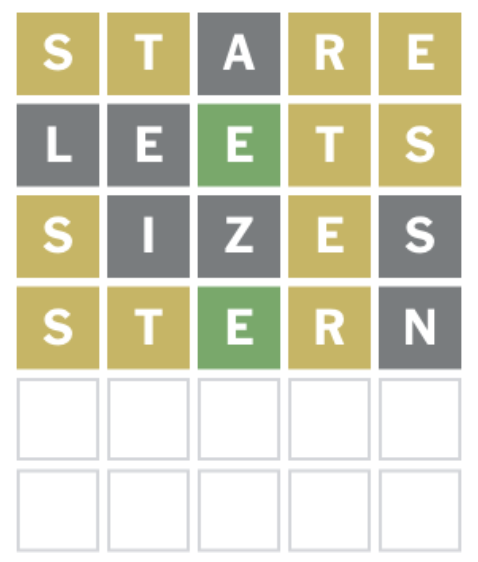}
\caption{\textbf{Visualization of Wordle game played by the CUA agent} – During a game session, the CUA agent is instructed to self-annotate or summarize the screen (via function calling). At the displayed state of the game, the CUA agent describes the screen as follows: ``\emph{The grid shows attempts with first guess (S T A R E) displaying two letters in yellow (S and T), 2nd guess (L E E T S) shows green for E, 3rd guess (S I Z E S) retained yellow for S and T, 4th guess (S T E R N) maintained yellow for S, T, and E.}''
The CUA agent's description indicates that it is unable to recognize correctly as well as consistently the color assigned to each letter while playing the game.}
\label{fig:wordle_example}
\end{figure}

Our analysis revealed that despite the CUA agent's proficiency in many complex tasks available on benchmarks, it exhibits consistent failures in correctly identifying and reasoning about colors in the context of the Wordle (see Fig.~\ref{fig:wordle_example} for an illustrative example). This discrepancy reinforces the significant doubts raised about current approach to AI (transformers) and their ability to achieve true general intelligence \citep{berglund2023reversal, wang2025reversal, petrov2025proof, gambardella2024language}.

\section{Background}

\subsection{Computer-Using Agent (CUA)}
OpenAI's Computer-Using Agents represents a key advancement in multimodal AI systems, enabling models to interact with computer interfaces in a human-like manner \citep{openai2025cua}:
\begin{quote}
``CUA builds on years of foundational research at the intersection of multimodal understanding and reasoning. By combining advanced GUI perception with structured problem-solving, it can break tasks into multi-step plans and adaptively self-correct when challenges arise. This capability marks the next step in AI development, allowing models
to use the same tools humans rely on daily and opening the door to a wide range of new applications.'' 
\end{quote}

Concretely, CUA operates through an iterative loop comprising of:
\begin{enumerate}
\item \textbf{Perception}: Incorporating screenshots into the model's context to provide a visual snapshot of the current computer state.
\item \textbf{Reasoning}: Utilizing chain-of-thought processes to determine subsequent actions, considering both current and prior interactions.
\item \textbf{Action}: Executing actions—such as coordinate-based clicking, scrolling, or typing—until a given task is completed or further input is required.
\end{enumerate}

This iterative architecture of CUA allows the model to handle complex, multi-step tasks, manage errors, and adapt to unforeseen changes. OpenAI's own offering of CUA is available as Operator \citep{openai2025operator}, which interfaces primarily as a browser using agent in the cloud.

Benchmarking efforts such as WebVoyager \citep{zhou2023webarena, he2024webvoyager}, OSWorld \citep{xie2024osworld}, and more recently BrowseComp \citep{openai2025browsecomp} have focused on performance evaluation of these agents in various complex scenarios and environments. However, these benchmarks offer only overall performance metrics, revealing little about specific limitations or their impact on task completion.

Although CUA is the model of focus in this study, other capable models with similar capabilities are available for development and use \citep{anthropic2024cua, qin2025ui, agashe2024agent,xu2024aguvis}. 
However, we believe the issues identified and conclusions made in this paper are not limited to CUA and should be broadly applicable to other multimodal models.

\subsection{Wordle Game Mechanics}
Wordle is a web-based word game developed by Josh Wardle and made available on the games section of The New York Times Company \citep{wordle2022}. The game play can be summarized into two key items: $(i)$ A player in a game has six attempts to guess a five-letter word, and $(ii)$ After each guess, the game provides feedback through color-coded tiles,  \textbf{\color{green}Green}: The letter is correct and in the correct position;  \textbf{\color{yellow}Yellow}: The letter is in the target word but in the wrong position; and   \textbf{\color{gray}Gray}: The letter is not in the target word.

Despite its apparent simplicity, Wordle presents a nuanced challenge that makes it an effective benchmark for evaluating the capabilities of today's computer-using agents. 

From an information theory perspective, Wordle serves as a practical application of concepts like entropy and information gain. Each guess partitions the set of possible solutions, and the feedback received reduces uncertainty about the target word. Optimal strategies aim to maximize expected information gain per guess, effectively minimizing the average number of attempts needed to solve the puzzle. This implies a clear reasoning path for an agent, one that is intelligent, to play and complete the game.

Although a large amount of data about Wordle is available on the internet, the terms of service of the game (\href{https://help.nytimes.com/hc/en-us/articles/115014893428-Terms-of-Service}{Wordle terms of service}) prevent bot scraping which introduces an interesting scenario in terms of generalization and out-of-distribution (OOD) testing for these models. Without loss of generality, one can assume that CUA is trained on the internet except scenarios that explicitly prevent scraping or are behind a paywall. Thus when encountering scenarios not present in their training data, the ability of the system to adapt and complete tasks will determine the level of reasoning and capability in these AI systems.

\subsection{Expected Capabilities for Wordle}
For an AI agent to successfully play Wordle, several fundamental capabilities are required:

\begin{itemize}
    \item \textbf{Visual Processing}: The ability to correctly identify letters and their corresponding color feedback in the game interface.
    \item \textbf{Logical Reasoning}: The ability to draw appropriate conclusions from color feedback (e.g., a green tiled letter is in the correct position).
    \item \textbf{Memory Integration}: The capacity to memorize information from previous guesses to inform future guesses.
    \item \textbf{Interface Manipulation}: The ability to use a virtual keyboard to type valid five-letter words and submit guesses.
    \item \textbf{Strategic Planning}: The capacity to select optimal guess words based on accumulated constraints.
\end{itemize}

Given CUA's capabilities in GUI interaction, visual processing, and language model capabilities, it is reasonable to expect the agent to perform well at this task. However, our findings suggest fundamental limitations in the model's perceptual and reasoning process.

\section{Experiments}

Our experiments are based on the CUA API made available by OpenAI and publicly available \href{https://platform.openai.com/docs/guides/tools-computer-use}{documentation} on the model. We did not have any additional model information. The design choices and instruction passed to the model with the API was thoroughly validated and evaluated for optimality through multiple experiments. The starting point of our evaluation system is the code provided by OpenAI available here \href{https://github.com/openai/openai-cua-sample-app}{Sample CUA App}.

\subsection{Wordle Performance Assessment}
We tasked the CUA agent to play Wordle with detailed system instruction and function to self-annotate its observation after each guess. The results presented here are based on the system prompt and observation tool definition presented in App.~\ref{app:prompt}. The assessment process is described in Alg.~\ref{alg:asses}.
\begin{algorithm}
\captionsetup{labelformat=empty}
\caption{Wordle Experiment Protocol}
\begin{algorithmic}[1]
\State Launch the New York Times Wordle website
\State Instruct CUA to play the game and explain its reasoning
\State Record the agent's performance, focusing on:
    \State \hspace{\algorithmicindent} Success/failure in reaching the solution
    \State \hspace{\algorithmicindent} Number of attempts required
    \State \hspace{\algorithmicindent} Accuracy of per attempt color recognition
    \State \hspace{\algorithmicindent} Quality of logical reasoning based on perceived colors
\end{algorithmic}
\label{alg:asses}
\end{algorithm}
We use the local playwright setting (chromium browser instance with start page set to Wordle) for all runs. Screenshots sent to the CUA model were screen grabs of the visible page sized at $1024\times768$.  The agent is instructed to complete the task with minimal additional input or confirmations from user to remove any human feedback. A single session (one play of Wordle) is considered complete if (a) the agent believes it has solved the Wordle, (b) the agent runs out of turns and the game ends, (c) A maximum number of API calls have been made ($60$ in our experiments), or  (d) the API times out or there was a network error. We perform $25$ runs \footnote{Additional runs were made in cases were there were multiple network related exits. The agent rarely ended up in the scenario of maximum API call based exits where it was stuck in a loop.} and report aggregated results over $8$ days of Wordle, totaling to $200$ runs of the CUA model on the game of Wordle per setting. 

\subsection{Wordle Performance}
The CUA agent demonstrated poor performance in Wordle, successfully solving the puzzle in only $5.36\%$ of cases, when it did manage to solve it, it required around $3$ guesses on average (see Table~\ref{tab:wordle_performance}). We will demonstrate one of the reasons why the model fails to succeed after three guesses, a point which falls within the early stages of the Wordle solving process.

Our analysis of the agent's reasoning revealed that color misidentification had a big impact in the observed performances. The subsequent sections are focusing on such an analysis.
\begin{table}[htp]
\centering
\caption{Summary of CUA performance in Wordle}
\label{tab:wordle_performance}
\begin{tabular}{lc}
\toprule
\textbf{Metric} & \textbf{CUA Agent}  \\
Avg. guesses per solved puzzle & 3.25 \\
Success rate & 5.36\% \\
\bottomrule
\end{tabular}
\end{table}

\begin{figure}[t]
    \centering
    \begin{subfigure}[b]{0.47\textwidth}
        \centering
        \includegraphics[width=\textwidth]{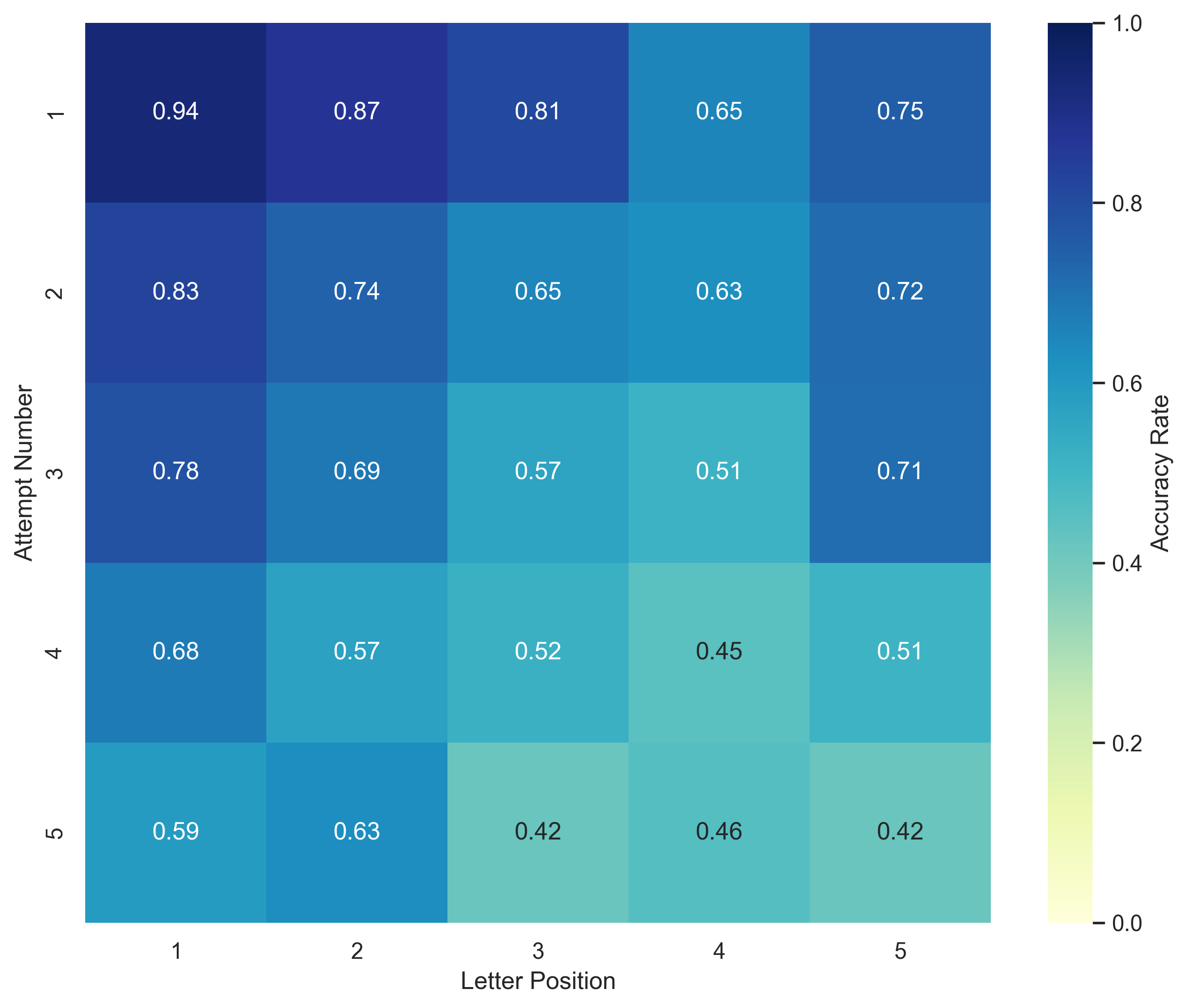}
        \label{fig:position_attempt_accuracy}
    \end{subfigure}
    \hfill
    \begin{subfigure}[b]{0.47\textwidth}
        \centering
        \includegraphics[width=0.68\textwidth]{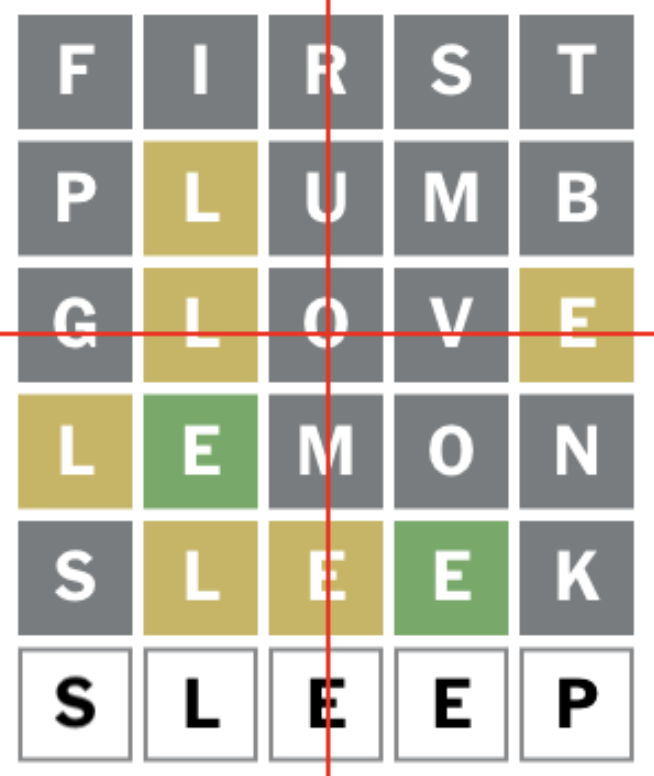}
        \vspace{.8cm}
    \end{subfigure}
    \caption{\textit{(Left)} \textbf{CUA color observation accuracy by letter position and attempt} - Heatmap showing color observation (as identified by the CUA model while playing the game) accuracy by letter position (1-5) and attempt number (1-5). Data shows highest accuracy (94\%) at position 1, attempt 1, with significant degradation in accuracy in later attempts and central positions. \textit{(Right)} \textbf{Depiction of the potential image tokenizer patch boundary} - In red we display the potential boundary of the image tokenization patches. We note that since we do not have access to the model and interact with the model via the provided API, the boundary has been plotted by deduction from the information gathered from OpenAI documentation. We believe that poor perception accuracy is the result of the image tokenization phenomena and the increasing complexity in terms of reasoning as the number of attempts increases.}
    \label{fig:combined-heatmap-patch}
\end{figure}
\subsection{Color Recognition by Position and Attempt}

Our findings revealed significant problems in CUA's color perception abilities across different letter positions and game attempt numbers. Figure \ref{fig:combined-heatmap-patch} (left) presents a heatmap showing the accuracy of color recognition for each letter position (1-5) across progressive game attempts (1-5)\footnote{The sixth attempt was removed as at times the screenshot was not timed appropriately by CUA to capture the Wordle feedback}.

From this heatmap we observe that letters in positions 1 and 5 (edges) showed mostly higher accuracy than positions 3 and 4 (center). Also, the recognition accuracy decreased substantially with each subsequent attempt. Finally, position $4$ showed the poorest performance overall.

These findings suggest that CUA's visual attention mechanisms struggle with maintaining consistent color perception across the entire Wordle grid, particularly as the game progresses and the grid becomes more populated with colored tiles.

Moreover, letters at the far right and left of the grid were identified more accurately relative to the letters on the center of the grid. One hypothesis that can explain this behavior is the tokenization of the input image by the CUA model. From model output metadata, we know that each screenshot image ($1024 \times 768$) is divided into $4$ patches in a $2\times2$ grid, each of size $512\times512$.  This would mean that the screenshot images with the Wordle grid are divided exactly around letter position 3 and attempt number 3, see Fig.~\ref{fig:combined-heatmap-patch} (right) for reference. 

\subsection{Model-Generated Color Observation Accuracy by Attempt}
To further understand the correlation between the color misidentification and the attempt number in the Wordle game, we propose in Fig.~\ref{fig:combined_plots_word_specific} (\textit{Left}) the per attempt average (across words position) color recognition accuracy rate. We clearly observe the decline in the agent's ability to correctly perceive all colors in a Wordle row as the game progresses through multiple attempts. The data shows a clear pattern of degradation from the first attempt ($42\%$ accuracy) to the fifth attempt ($6\%$ accuracy). This provides compelling evidence of how the CUA's perceptual abilities break down with increasing number of turns or reasoning complexity.

\begin{figure}[t]
    \centering
    \begin{subfigure}[b]{0.48\textwidth}
        \centering
        \includegraphics[width=1\textwidth]{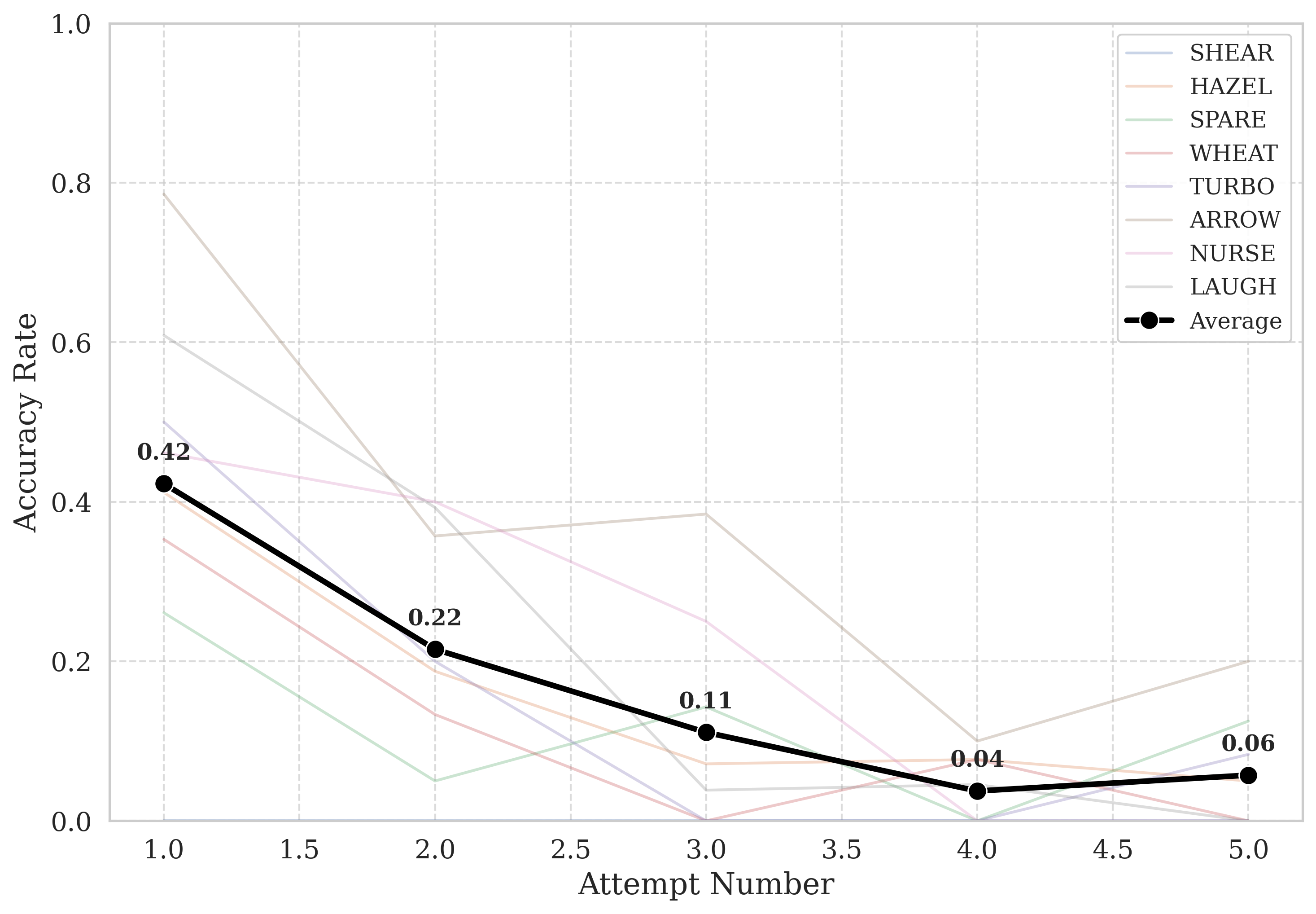}
    \end{subfigure}
    \hfill
    \begin{subfigure}[b]{0.48\textwidth}
        \centering
\includegraphics[width=1.0\textwidth]{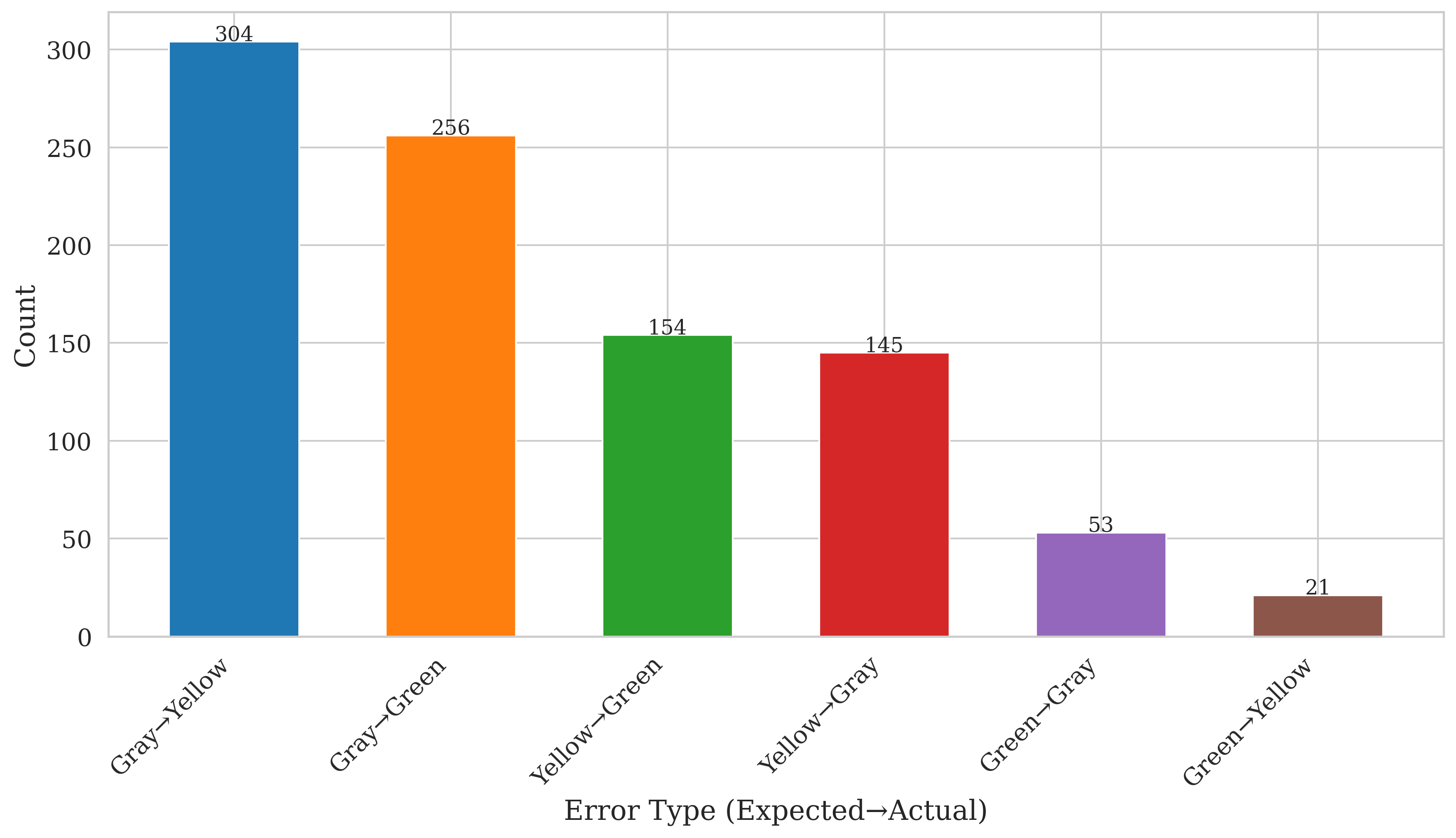}
    \end{subfigure}
    \caption{(\textit{Left}) \textbf{Model-generated color observation accuracy by attempt number}. The average accuracy (bold black line) decreases dramatically from $42\%$ in attempt $1$ to $6\%$ in attempt $5$, with individual words showing varying decline patterns. This demonstrates the agent's increasing difficulty in correctly perceiving colors as the game progresses. Note that the accuracy in the first attempt, although relatively higher than the rest, indicates that there is a fundamental perception problem in the model. (\textit{Right}) \textbf{Most common observation errors by color type} - Bar chart showing the frequency of different types of color recognition errors. Here, \emph{Expected} is the color that the model should have observed while \emph{Actual} is the color observation made by the model. Gray→Yellow and Gray→Green were the most common errors, followed by Yellow→Green and Yellow→Gray. This suggests the agent has particular difficulty distinguishing gray tiles from colored tiles. Alternatively, the model might just be biased toward \emph{seeing} specific colors even when gray (higher confidence in its own chain of thought and guesses \citep{chowdhury2025truthfulness}).}
    \label{fig:combined_accuracy_color_mix}
  \end{figure}

The decline suggests that as more colored tiles populate the grid, the agent's visual processing system becomes increasingly unreliable, with accuracy dropping to near zeros by the fourth and fifth attempts.

\subsection{Most Common Observation Errors by Color Type}
\begin{figure}[t]
    \centering
    \begin{subfigure}[b]{0.48\textwidth}
        \centering
        \includegraphics[width=\textwidth]{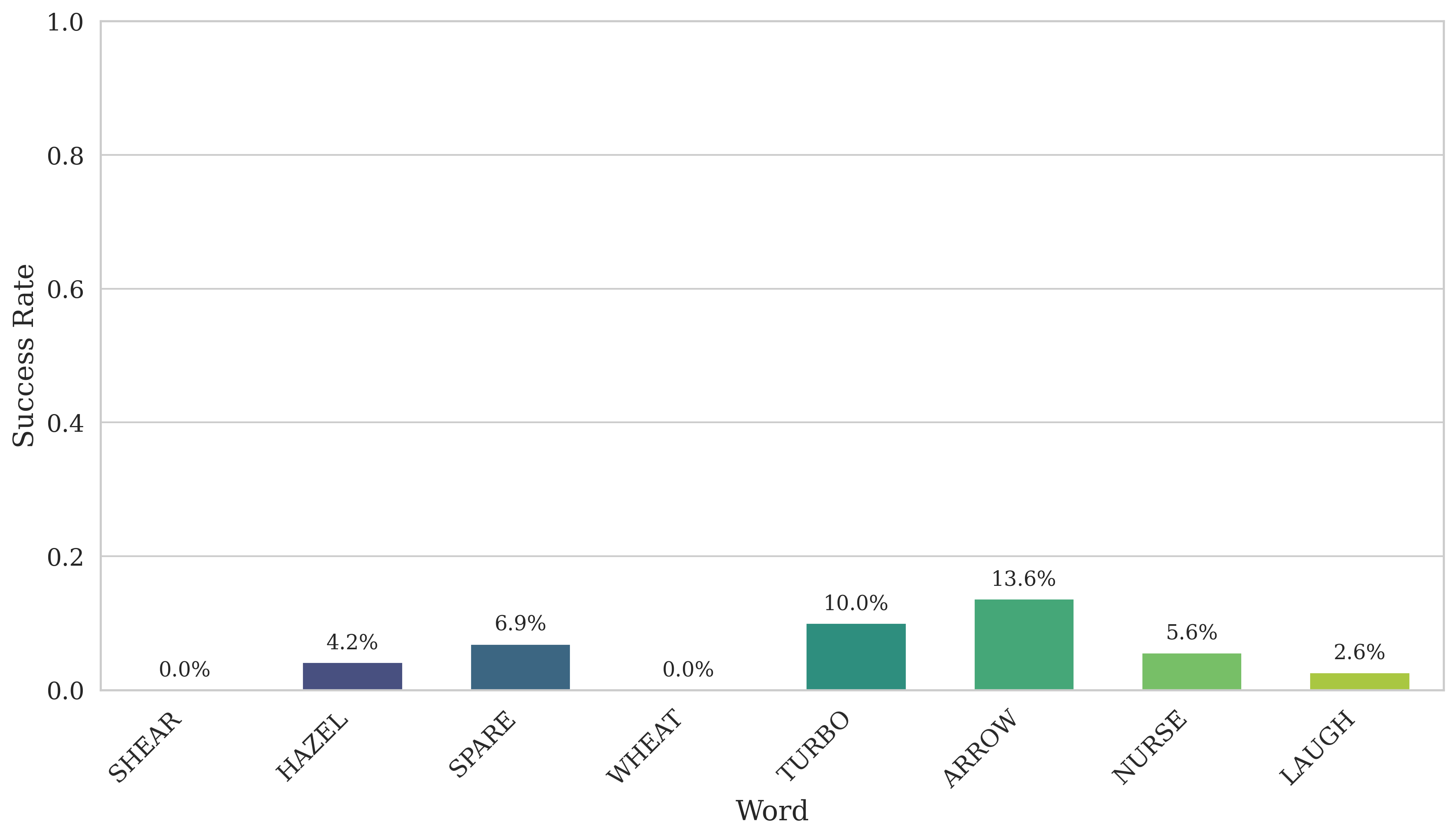}
    \end{subfigure}
    \hfill
    \begin{subfigure}[b]{0.48\textwidth}
        \centering
        \includegraphics[width=\textwidth]{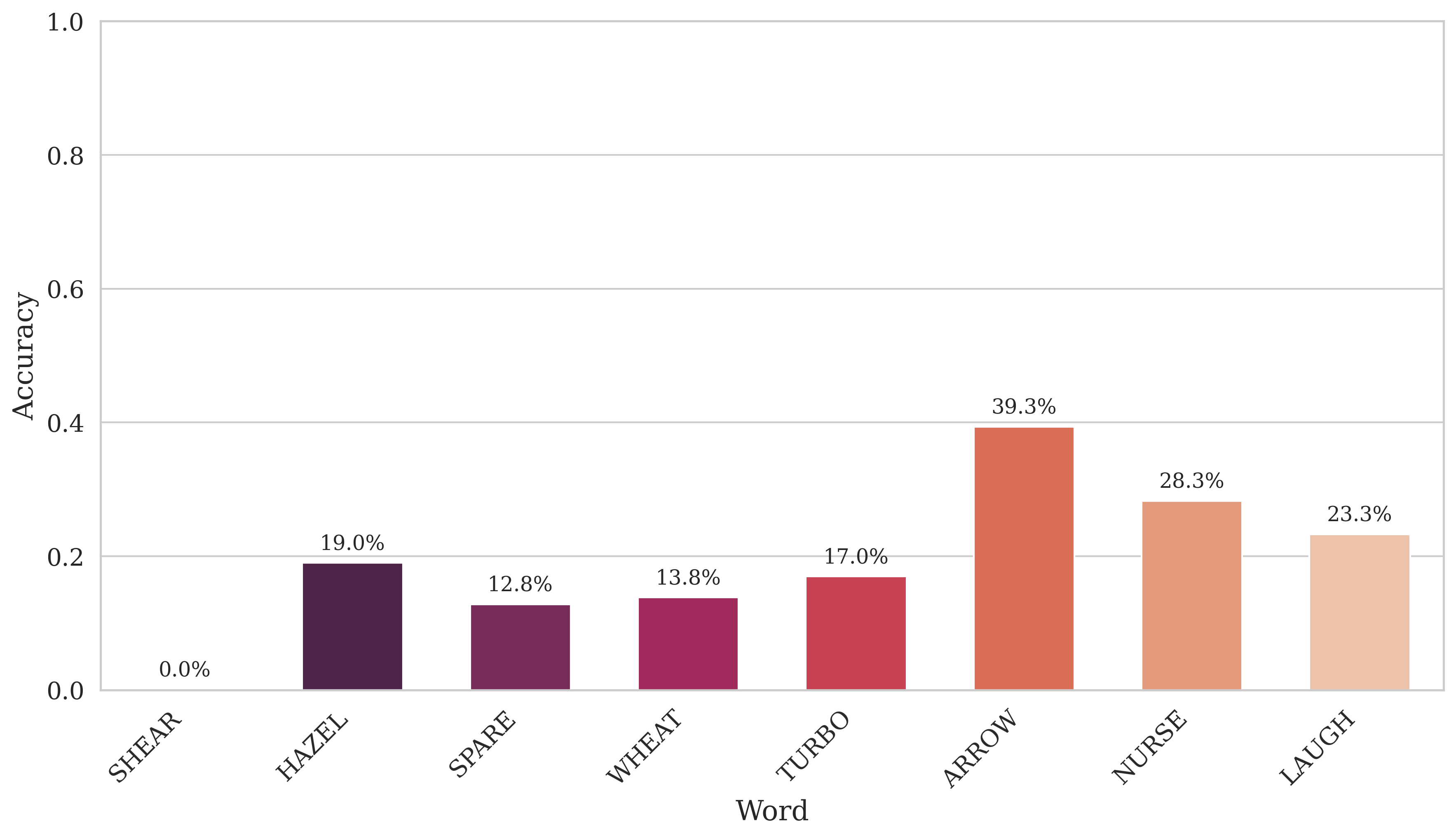}
    \end{subfigure}
    \caption{\textit{(Left)} \textbf{Success rate by word} - Bar chart showing success rate by target Wordle word. ARROW ($13.6\%$) and TURBO ($10.0\%$) had the highest success rates, while SHEAR and WHEAT had a $0\%$ success rate. The pattern closely mirrors observation accuracy, supporting the connection between color perception and game performance. \textit{(Right)} \textbf{Model-generated color observation accuracy (average over attempts)} - Bar chart showing model-generated observation accuracy by word. ARROW shows the highest accuracy ($39.3\%$), followed by NURSE ($28.3\%$) and LAUGH ($23.3\%$), while SHEAR shows $0\%$ accuracy. This suggests word-specific factors do influence color perception ability.}
    \label{fig:combined_plots_word_specific}
\end{figure}

We present in this section analysis on the misidentifications, to reveal any specific patterns in the CUA's perception capabilities. Figure \ref{fig:combined_plots_word_specific} (\textit{Right}) presents a comprehensive breakdown of color recognition errors by color type.

The predominance of Gray→Yellow and Gray→Green errors reveals a systematic bias in the CUA's color perception: it tends to \emph{hallucinate} colors on gray tiles more often than it fails to perceive actual colors. This asymmetric error pattern suggests fundamental issues in how the model processes and interprets color information in the context of the Wordle game interface. It should be noted that often in our experiment, the model believes it successfully solve the Wordle game because of hallucinating green tiles for all letters. It is interesting to note that this optimistic bias is common with LLM and often due to their RLHF post-training \citep{sharma2023towards, leng2024taming, kadavath2022language, achiam2023gpt}. 

\subsection{Word-Specific Performance Analysis}
Additional analysis revealed significant variation in CUA's performance across different target words. The plots in  Fig.~\ref{fig:combined_plots_word_specific} present the success rate and color observation accuracy by word. 

\begin{figure}[h]
\centering
\includegraphics[width=0.5\textwidth]{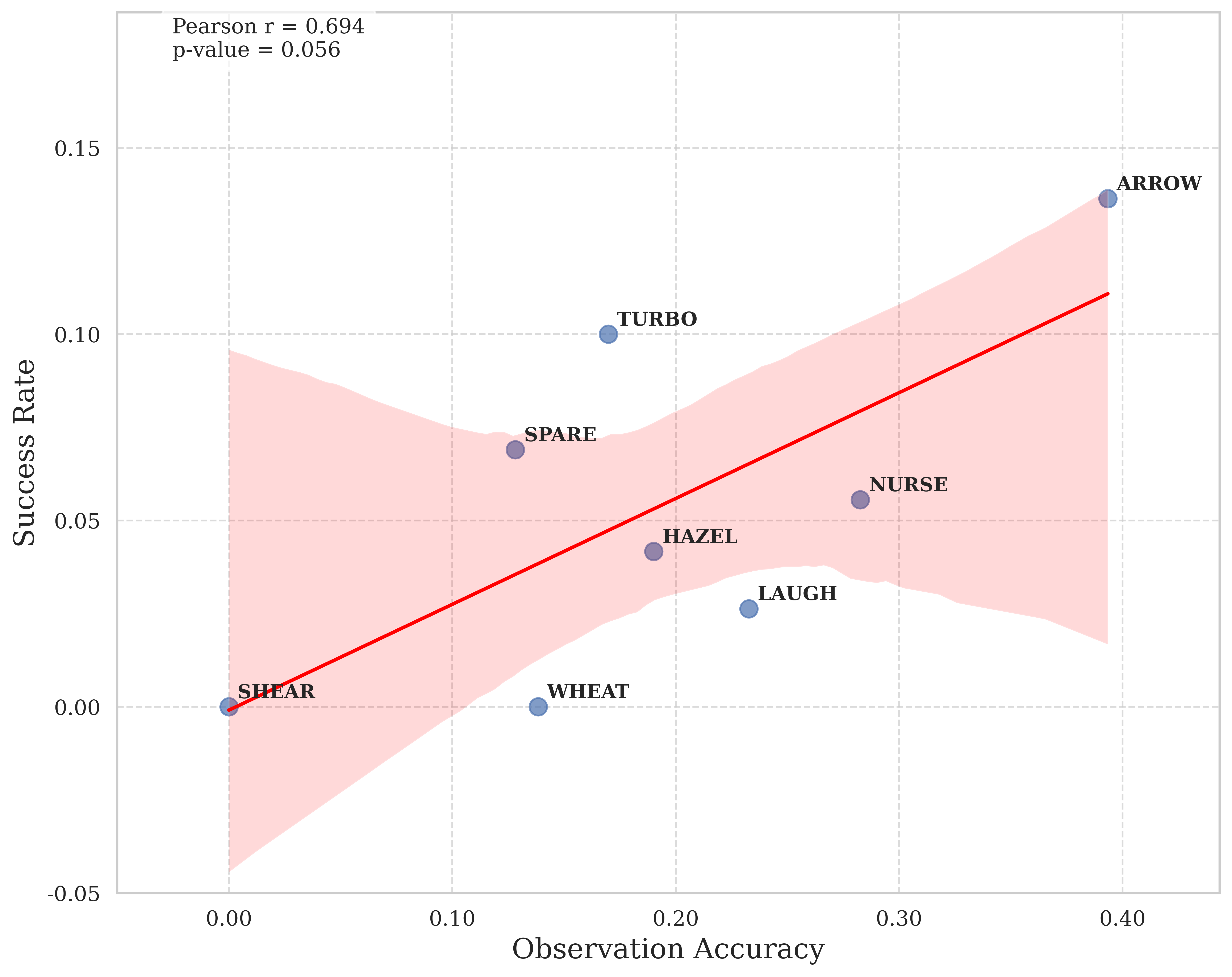}
\caption{\textbf{Correlation between observation color accuracy and word success rate} - Scatter plot showing the correlation between observation accuracy and success rate across different words. With a Pearson r of $0.694$ and p-value of $0.056$, there is a strong positive correlation between the agent's ability to correctly perceive colors and its ability to solve the Wordle puzzle, highlighting how perceptual failures directly impact task performance.}
\label{fig:correlation}
\end{figure}

Fig.~\ref{fig:correlation} presents the correlation between observation color accuracy and word success rate. This result shows that the factors are indeed correlated with a Pearson correlation coefficient of $0.694$ (p-value = $0.056$). Though not surprising, as not being able to perceive the current state of the game well should mean the chances of winning the game are minimized, the experiments do present a strong validation on the hypothesis.

\section{Discussion}
In this section, we summarize our thoughts on computer using models and present a discussion on current limitations and future directions for both research and development with these systems. 

The contrast between CUA's benchmark performance and its struggles with the Wordle game highlights one of the challenges AI models are currently facing. Architecturally, while tokenizers with transformers have allowed us to achieve major improvements in AI systems, they do posses some fundamental limitations. The predominant approach to solving these challenges have been through reinforcement training of the models with more data. 

The experiments in this paper brings back the question of meaningful generalization vs memorization. The Wordle game always starts with a pop-up about terms and condition which the model had no difficult in completing. CUA was even able to follow up by clicking \emph{Play} and then closing the game rules dialog that followed. However, when it came to playing the actual game, the model performance began deteriorating pretty quickly.

Our quantitative analysis particularly underscores this gap. The dramatic decline in color recognition accuracy across successive attempts (from $42\%$ in the first attempt to $6\%$ by the fifth) reveals a fundamental brittleness in CUA's perceptual systems.

All these lead us to believe that the system while demonstrating useful capabilities are still far away from solving tasks that require non-trivial planning, perception, and reasoning. 

\subsection{Potential Causes}
Without access to the model architecture, we can only hypothesize about the underlying causes of this context-dependent color recognition failure.

 \paragraph{Training Data:} It is reasonable to assume that during training the model was exposed to examples of color recognition tasks, several articles on the game of Wordle, as well as planning strategies of how to play and win games. However, we believe that  the model was not explicitly trained on Wordle. The New York Times \href{https://help.nytimes.com/hc/en-us/articles/115014893428-Terms-of-Service}{terms of service} clearly outline against the use of its content for training AI models. This makes the actual task of playing Wordle, a challenging task for the model. 
\paragraph{Non-localized Perception:} In visual environments like Wordle, the model's attention may be diffused across multiple elements (letters, colors, grid structure, prompt, tools), leading to less precise color perception \citep{wolfe2020visual}. In fact, our tests to perform color recognition on simple color grids as well as identifying numbers using Ishihara test revealed the model was indeed capable of completing and succeeding on these tests that are not requiring multiple steps of reasoning.
\paragraph{Spatial Processing and Tokenization:} The systematic pattern of higher accuracy at edge positions (positions 1 and 5) compared to central positions suggests potential limitations in how CUA processes images. This might indicate some issues induced by discretization of such a continuous space \citep{wang2023makes}. 

Our analysis is limited by the fact that we do not have access to the model nor to its training data, and therefore can not do in depth analysis of the underlying problem. However, it is clear that the problem we are facing here is not isolated.  
Interestingly, OpenAI released a solution for this limitation in different a series of their models (\emph{o3, o4-mini}) at the time of our writing dubbed as \href{https://openai.com/index/thinking-with-images/}{thinking with images}. Their proposed solution relies on using additional tools to zoom-in, crop, and process different parts of an input image to gather details and reason. Note that this approach was similarly proposed in recent articles \citep{liu2023visual, wu2023v}. 
We believe this approach circumvents the problem of tokenization process in images and will require a more fundamental approach to truly overcome.

\section{Conclusion}

Our investigation reveals a significant limitation in the CUA agent's ability to process color information and reasoning to succeed in Wordle. The  difference in performance between isolated color recognition tests and the Wordle game highlights the complexity of visual processing in multimodal AI systems. Besides it confirms the challenges induced by multi-step tasks that are (potentially) out of domain.

The gap between the rhetoric surrounding AI agents and their actual performance suggests that achieving AGI requires more fundamental improvements beyond current algorithms, be it tokenization, architecture, training, or reward modeling. While systems like CUA do demonstrate impressive capabilities, their clear limitations open up new avenues for research and benchmarking.
 
We hope that 
future work will explore robust methods for enhancing context-dependent perception, including specialized training and architectural modifications to better support agentic capabilities in unseen multi-step environments. 


\bibliographystyle{apalike}
\bibliography{main}

\begin{thebibliography}{}

\bibitem[Achiam et~al., 2023]{achiam2023gpt}
Achiam, J., Adler, S., Agarwal, S., Ahmad, L., Akkaya, I., Aleman, F.~L., Almeida, D., Altenschmidt, J., Altman, S., Anadkat, S., et~al. (2023).
\newblock Gpt-4 technical report.
\newblock {\em arXiv preprint arXiv:2303.08774}.

\bibitem[Agashe et~al., 2024]{agashe2024agent}
Agashe, S., Han, J., Gan, S., Yang, J., Li, A., and Wang, X.~E. (2024).
\newblock Agent s: An open agentic framework that uses computers like a human.
\newblock {\em arXiv preprint arXiv:2410.08164}.

\bibitem[Anthropic, 2024]{anthropic2024cua}
Anthropic (2024).
\newblock Introducing computer use.
\newblock Anthropic Blog.
\newblock \url{https://www.anthropic.com/news/3-5-models-and-computer-use}.

\bibitem[Benveniste, 2022]{wordle2022}
Benveniste, A. (2022).
\newblock The sudden rise of wordle.
\newblock {The New York Times}.
\newblock \url{https://www.nytimes.com/2022/01/31/crosswords/nyt-wordle-purchase.html}.

\bibitem[Berglund et~al., 2023]{berglund2023reversal}
Berglund, L., Tong, M., Kaufmann, M., Balesni, M., Stickland, A.~C., Korbak, T., and Evans, O. (2023).
\newblock The reversal curse: Llms trained on" a is b" fail to learn" b is a".
\newblock {\em arXiv preprint arXiv:2309.12288}.

\bibitem[Chen et~al., 2025]{chen2025janus}
Chen, X., Wu, Z., Liu, X., Pan, Z., Liu, W., Xie, Z., Yu, X., and Ruan, C. (2025).
\newblock Janus-pro: Unified multimodal understanding and generation with data and model scaling.
\newblock {\em arXiv preprint arXiv:2501.17811}.

\bibitem[Chowdhury et~al., 2025]{chowdhury2025truthfulness}
Chowdhury, N., Johnson, D., Huang, V., Steinhardt, J., and Schwettmann, S. (2025).
\newblock Investigating truthfulness issues in a pre-release o3 model.
\newblock \url{https://transluce.org/investigating-o3-truthfulness}.

\bibitem[Gambardella et~al., 2024]{gambardella2024language}
Gambardella, A., Iwasawa, Y., and Matsuo, Y. (2024).
\newblock Language models do hard arithmetic tasks easily and hardly do easy arithmetic tasks.
\newblock {\em arXiv preprint arXiv:2406.02356}.

\bibitem[He et~al., 2024]{he2024webvoyager}
He, H., Yao, W., Ma, K., Yu, W., Dai, Y., Zhang, H., Lan, Z., and Yu, D. (2024).
\newblock Webvoyager: Building an end-to-end web agent with large multimodal models.
\newblock {\em arXiv preprint arXiv:2401.13919}.

\bibitem[Kadavath et~al., 2022]{kadavath2022language}
Kadavath, S., Conerly, T., Askell, A., Henighan, T., Drain, D., Perez, E., Schiefer, N., Hatfield-Dodds, Z., DasSarma, N., Tran-Johnson, E., et~al. (2022).
\newblock Language models (mostly) know what they know.
\newblock {\em arXiv preprint arXiv:2207.05221}.

\bibitem[Leng et~al., 2024]{leng2024taming}
Leng, J., Huang, C., Zhu, B., and Huang, J. (2024).
\newblock Taming overconfidence in llms: Reward calibration in rlhf.
\newblock {\em arXiv preprint arXiv:2410.09724}.

\bibitem[Liu et~al., 2023]{liu2023visual}
Liu, H., Li, C., Wu, Q., and Lee, Y.~J. (2023).
\newblock Visual instruction tuning.
\newblock {\em Advances in neural information processing systems}, 36:34892--34916.

\bibitem[MetaAI, 2025]{llama2024}
MetaAI (2025).
\newblock Llama 4 models.
\newblock Llama Blog.
\newblock \url{https://www.llama.com/models/llama-4/}.

\bibitem[MistralAI, 2024]{pixtral2024}
MistralAI (2024).
\newblock Pixtral large.
\newblock Mistral Blog.
\newblock \url{https://mistral.ai/news/pixtral-large}.

\bibitem[{OpenAI}, 2025a]{openai2025browsecomp}
{OpenAI} (2025a).
\newblock Browsecomp: a benchmark for browsing agents.
\newblock OpenAI Blog.
\newblock \url{https://openai.com/index/browsecomp/}.

\bibitem[{OpenAI}, 2025b]{openai2025cua}
{OpenAI} (2025b).
\newblock Computer-using agent.
\newblock OpenAI Blog.
\newblock \url{https://openai.com/index/computer-using-agent/}.

\bibitem[{OpenAI}, 2025c]{openai2025operator}
{OpenAI} (2025c).
\newblock Introducing operator: Our first ai agent that can use computers.
\newblock OpenAI Blog.
\newblock \url{https://openai.com/blog/introducing-operator}.

\bibitem[Petrov et~al., 2025]{petrov2025proof}
Petrov, I., Dekoninck, J., Baltadzhiev, L., Drencheva, M., Minchev, K., Balunovi{\'c}, M., Jovanovi{\'c}, N., and Vechev, M. (2025).
\newblock Proof or bluff? evaluating llms on 2025 usa math olympiad.
\newblock {\em arXiv preprint arXiv:2503.21934}.

\bibitem[Qin et~al., 2025]{qin2025ui}
Qin, Y., Ye, Y., Fang, J., Wang, H., Liang, S., Tian, S., Zhang, J., Li, J., Li, Y., Huang, S., et~al. (2025).
\newblock Ui-tars: Pioneering automated gui interaction with native agents.
\newblock {\em arXiv preprint arXiv:2501.12326}.

\bibitem[Sharma et~al., 2023]{sharma2023towards}
Sharma, M., Tong, M., Korbak, T., Duvenaud, D., Askell, A., Bowman, S.~R., Cheng, N., Durmus, E., Hatfield-Dodds, Z., Johnston, S.~R., et~al. (2023).
\newblock Towards understanding sycophancy in language models.
\newblock {\em arXiv preprint arXiv:2310.13548}.

\bibitem[Wang and Sun, 2025]{wang2025reversal}
Wang, B. and Sun, H. (2025).
\newblock Is the reversal curse a binding problem? uncovering limitations of transformers from a basic generalization failure.
\newblock {\em arXiv preprint arXiv:2504.01928}.

\bibitem[Wang et~al., 2023]{wang2023makes}
Wang, G., Ge, Y., Ding, X., Kankanhalli, M., and Shan, Y. (2023).
\newblock What makes for good visual tokenizers for large language models?
\newblock {\em arXiv preprint arXiv:2305.12223}.

\bibitem[Wolfe, 2020]{wolfe2020visual}
Wolfe, J.~M. (2020).
\newblock Visual search: How do we find what we are looking for?
\newblock {\em Annual review of vision science}, 6(1):539--562.

\bibitem[Wu and Xie, 2023]{wu2023v}
Wu, P. and Xie, S. (2023).
\newblock {V*: Guided Visual Search as a core mechanism in multimodal LLMs.}
\newblock In {\em CVF Conference on Computer Vision and Pattern Recognition (CVPR)}.

\bibitem[Xie et~al., 2024]{xie2024osworld}
Xie, T., Zhang, D., Chen, J., Li, X., Zhao, S., Cao, R., Hua, T.~J., Cheng, Z., Shin, D., Lei, F., et~al. (2024).
\newblock Osworld: Benchmarking multimodal agents for open-ended tasks in real computer environments.
\newblock {\em Advances in Neural Information Processing Systems}, 37:52040--52094.

\bibitem[Xu et~al., 2024]{xu2024aguvis}
Xu, Y., Wang, Z., Wang, J., Lu, D., Xie, T., Saha, A., Sahoo, D., Yu, T., and Xiong, C. (2024).
\newblock Aguvis: Unified pure vision agents for autonomous gui interaction.
\newblock {\em arXiv preprint arXiv:2412.04454}.

\bibitem[Zhou et~al., 2023]{zhou2023webarena}
Zhou, S., Xu, F.~F., Zhu, H., Zhou, X., Lo, R., Sridhar, A., Cheng, X., Ou, T., Bisk, Y., Fried, D., et~al. (2023).
\newblock Webarena: A realistic web environment for building autonomous agents.
\newblock {\em arXiv preprint arXiv:2307.13854}.

\end{thebibliography}

\newpage

\appendix
\section{Prompts and Tools}
\label{app:prompt}

\subsection*{System Prompt}
\begin{lstlisting}
# Objective
You are tasked with playing the game **Wordle**, with the goal of guessing the hidden *5 letter word* in a maximum of *six attempts*. Each time a guess is made, the color of the tiles will change to indicate how close the guess is to the target word. Based on this feedback, you will make subsequent guesses until you have guessed the target word.

The colors of the tiles are as follows:
- **Letter is in the target word and in the correct position** - Marked with green tile.
- **Letter is in the target word but in the wrong position** - Marked with yellow tile.
- **Letter is not in the target word** - Marked with black (gray) tile.

# Instructions for Playing the Game:
1. `type` to provide your 5 letter guess.
2. Use `keypress` ENTER to submit your guess. The guess is not submitted until keypress ENTER.
3. Use `screenshot` to take a screenshot of the current state of the game.
4. Use `update_wordle_game_state` to save the feedback obtained from your guess (last row with letters in the Wordle grid). 
5. Repeat steps 1 to 4 until you have guessed the target word (All letters are green) or you have made 6 attempts.
\end{lstlisting}

\subsection*{Tool Definition}
\begin{lstlisting}
"name": "update_wordle_game_state",
"description": """This tool records the observed information from the screenshot about letters, their positions and their absence.
    ## Example
    - Guess: "PLATE"
    - Observation: "GYBBB"
        - 'P' is correctly placed at position 1 ('G' - Green).
        - 'L' is in the word but not in position 2 ('Y' - Yellow).
        - 'A', 'N', and 'E' are not in the word ('B' - Black).
    """,
"parameters": {
    "type": "object",
    "properties": {
        "summary": {
            "type": "string",
            "description": "Use this argument to summarize the information observed in the screenshot. Record your thinking and carefully analyze the screenshot to come up with the letters and color observation for the most recent guess.",
        },
        "letters": {
            "type": "string",
            "description": "The guessed word. Each letter corresponds to a position in the word.",
            "minLength": 5,
            "maxLength": 5,
        },
        "observation": {
            "type": "string",
            "description": "A string of length 5 representing the feedback for each letter in the guessed word. Use 'G' for green (correct position), 'Y' for yellow (correct letter, wrong position), and 'B' for black (letter not in target word).",
            "minLength": 5,
            "maxLength": 5,
        },
    }
\end{lstlisting}

\end{document}